\documentclass[letterpaper, 10 pt, conference]{ieeeconf}  

\IEEEoverridecommandlockouts                              

\usepackage{graphicx}
\usepackage{amsmath}
\usepackage{amssymb}
\usepackage{booktabs}
\usepackage{float}
\usepackage{cuted}
\usepackage{capt-of}
\usepackage{xcolor}
\usepackage{enumerate}
\usepackage{tabularx}
\usepackage{subcaption}

\def\etal{\textit{et al}.}

\title{\LARGE \bf
Video Waterdrop Removal via Spatio-Temporal Fusion in Driving Scenes
}

\author{Qiang Wen, Yue Wu, and Qifeng Chen
\thanks{Qiang Wen (qwenab@connect.ust.hk), Yue Wu (ywudg@connect.ust.hk), and  Qifeng Chen (cqf@ust.hk) are with the Department of Computer Science and Engineering, HKUST.}
}

\begin{document}

\maketitle
\thispagestyle{empty}
\pagestyle{empty}

\begin{abstract}

  The waterdrops on windshields during driving can cause severe visual obstructions, which may lead to car accidents. Meanwhile, the waterdrops can also degrade the performance of a computer vision system in autonomous driving. To address these issues, we propose an attention-based framework that fuses the spatio-temporal representations from multiple frames to restore visual information occluded by waterdrops. Due to the lack of training data for video waterdrop removal, we propose a large-scale synthetic dataset with simulated waterdrops in complex driving scenes on rainy days. To improve the generality of our proposed method, we adopt a cross-modality training strategy that combines synthetic videos and real-world images. Extensive experiments show that our proposed method can generalize well and achieve the best waterdrop removal performance in complex real-world driving scenes.

\end{abstract}

\vspace{-3mm}
\section{INTRODUCTION}

With the increasing applications of visual perception methods~\cite{bruls2018mark,valada2017adapnet,kumar2021omnidet,zou2022real} in robotics and autonomous driving, the robustness of these methods has become more and more important in their visual system. However, the performance of current vision methods significantly degrades in rainy weather in autonomous driving scenarios, since raindrops on the windshield or camera lens cause inevitable visual obstruction, as mentioned in Porav et al.~\cite{porav2019can}. Therefore, removing waterdrops from videos on rainy days is highly important for self-driving cars and various robot applications.

\begin{figure}[t!]
\centering
\begin{tabular}{@{}c@{\hspace{1mm}}c@{\hspace{1mm}}c@{}}
\includegraphics[width=0.16\textwidth]{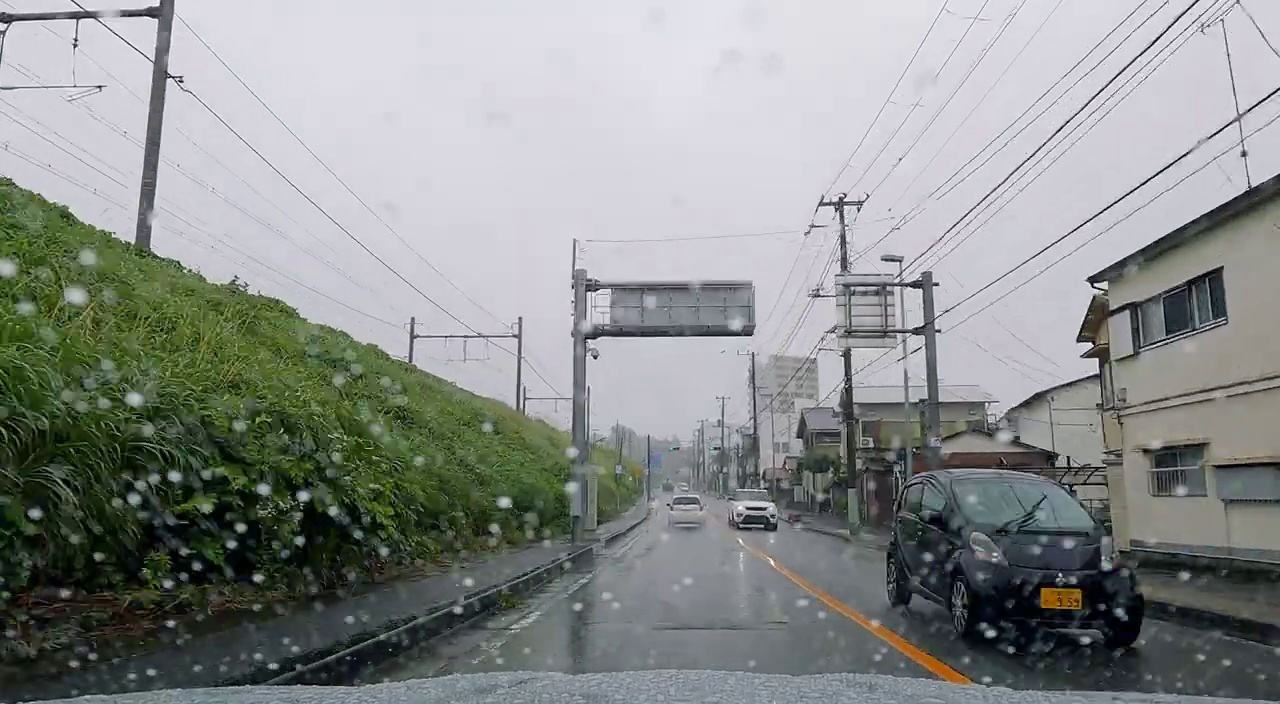}     & \includegraphics[width=0.16\textwidth]{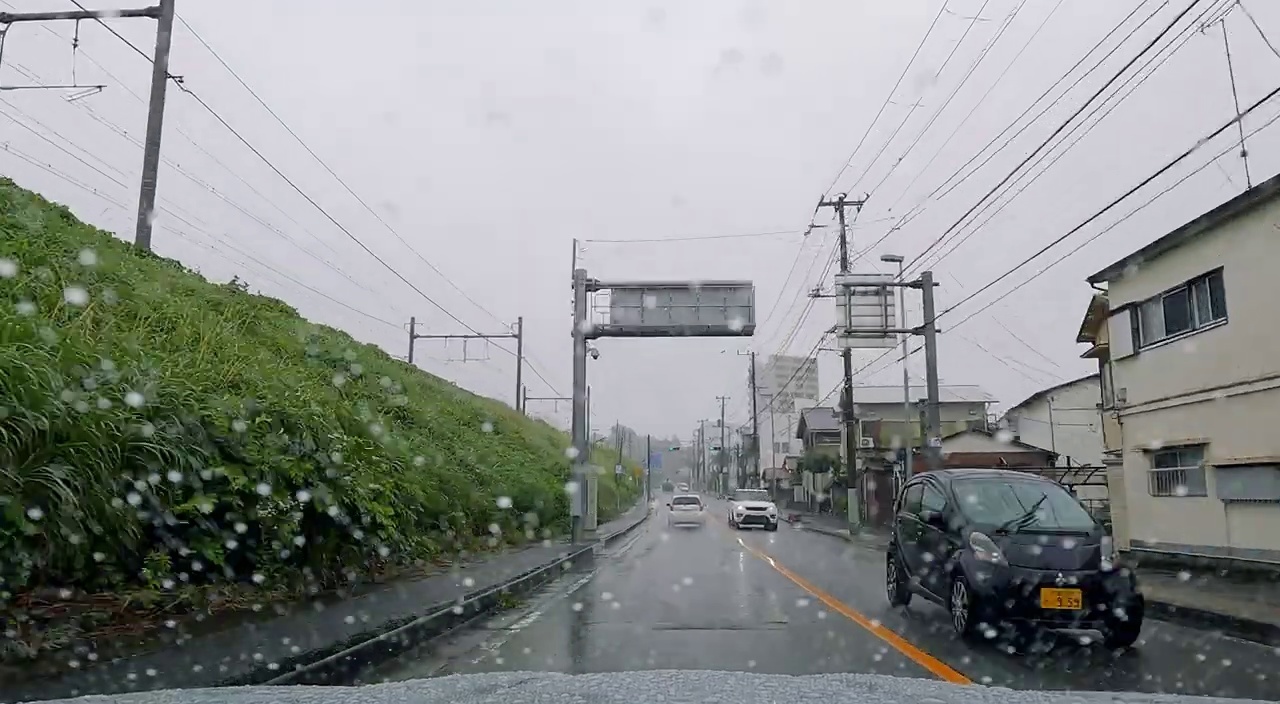}     & \includegraphics[width=0.16\textwidth]{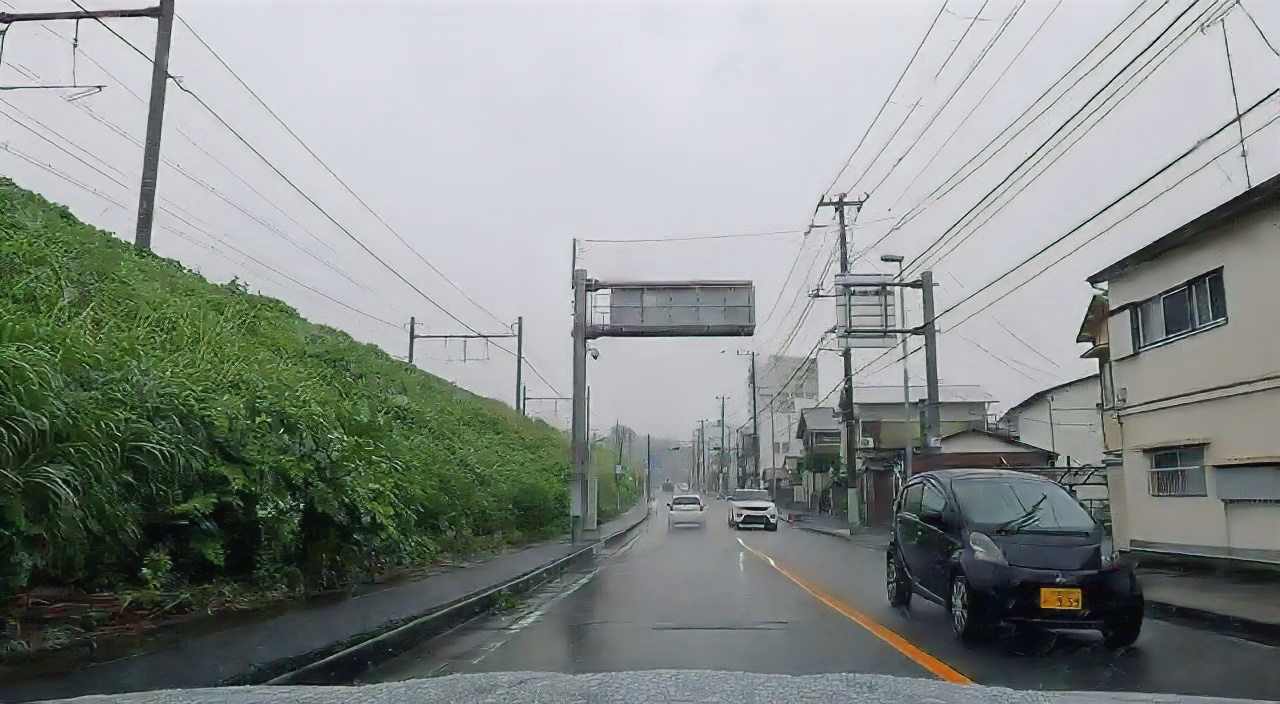} \\
Input (sparse case) & Restormer~\cite{Zamir2021Restormer} & Our Result\\
\includegraphics[width=0.16\textwidth]{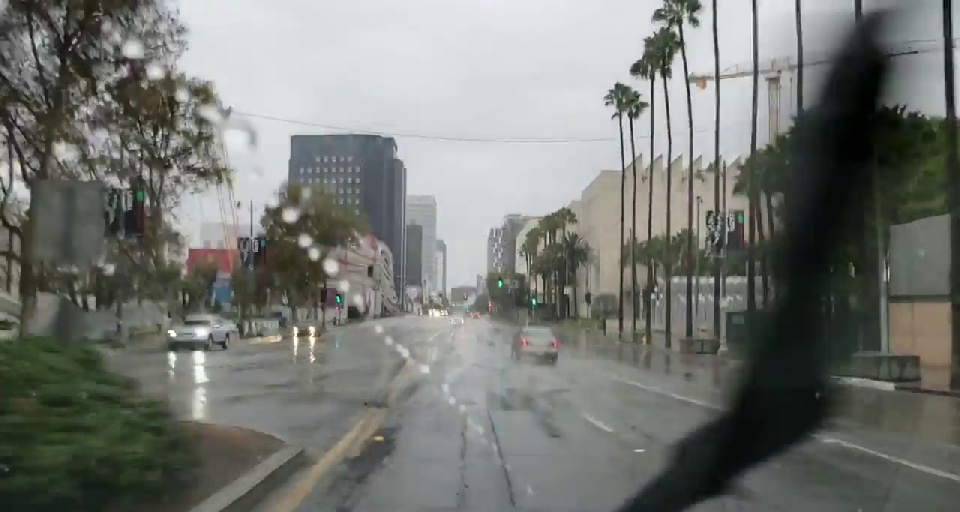} & \includegraphics[width=0.16\textwidth]{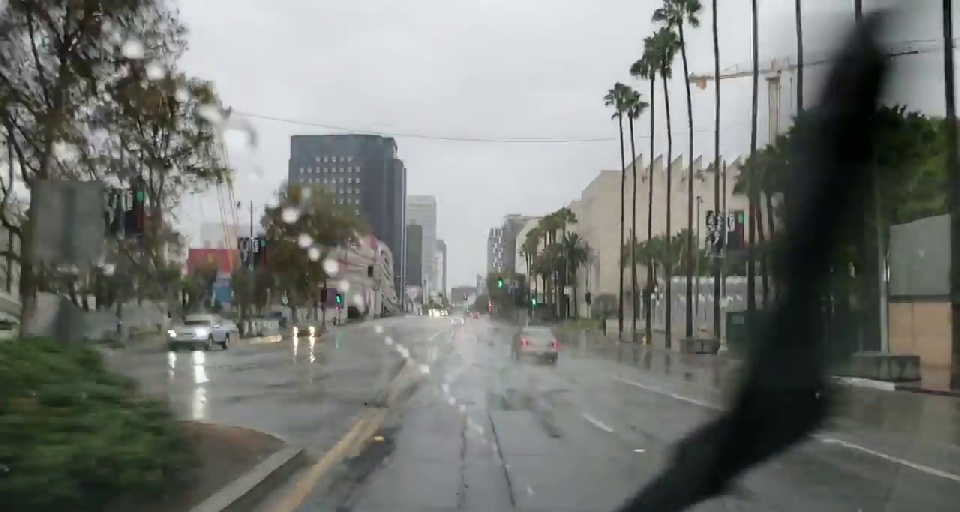} & \includegraphics[width=0.16\textwidth]{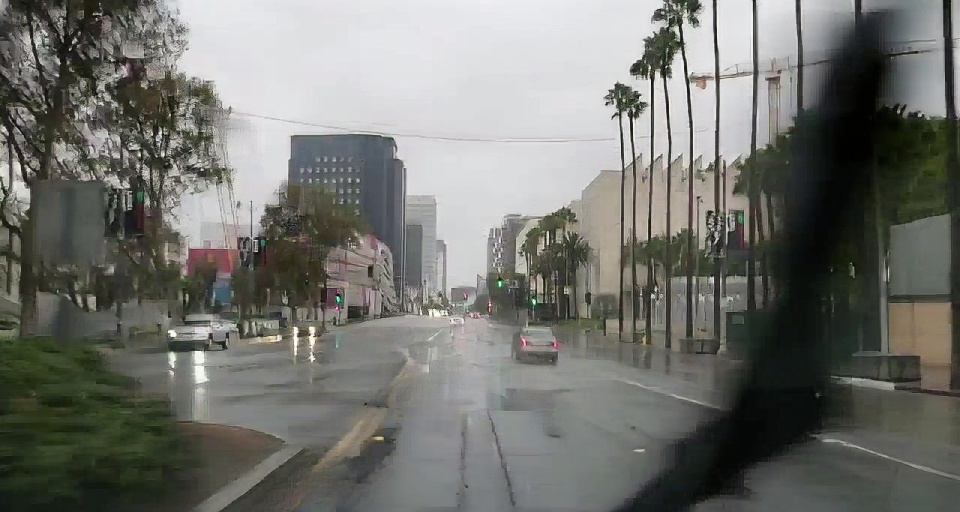}\\
Input (streak case) & Restormer~\cite{Zamir2021Restormer} & Our Result\\
\end{tabular}
\vspace{-2mm}    
\caption{Real-world driving videos with waterdrops.~\label{fig:1}}
\vspace{-6mm}
\end{figure}

Although many researchers~\cite{fu2017clearing,zhang2019image,chen2021hinet,li2019heavy,chen2019gated,wang2020model,yi2021structure,zamir2021multi,chen2021robust,wang2021rain,chen2020pmhld,wang2020model,jiang2020multi} propose dedicated frameworks to remove rain streaks, video waterdrop removal receive much less attention and is still an open question. Since there exists a significant geometric gap between waterdrops and rain streaks, the dedicated methods proposed to remove rain streaks cannot perform well on the waterdrop removal task as mentioned in~\cite{quan2021removing}. As shown in Fig.~\ref{fig:1}, directly applying a state-of-the-art rain streak removal method~\cite{Zamir2021Restormer} to remove waterdrops will not yield satisfactory results. Instead of being like the line shape of rain streaks, waterdrops tend to be ellipse scattered on windshields. And each waterdrop usually occupies a larger area than a rain streak, thus being more difficult to deal with.


To tackle the waterdrop removal problem, some researchers~\cite{eigen2013restoring,qian2018attentive,quan2019deep,shi2021stereo,hao2019learning,porav2019can} propose specialized frameworks to remove waterdrops from a single image. However, they still cannot handle complex driving videos, which are very common in Autonomous Driving. On the one hand, their datasets are usually collected by sprinkling waterdrops on the glass, thus there is a large domain gap between these data and real driving scenes. On the other hand, their methods lack temporal information utilization, which is necessary for video tasks.

The lack of paired real-world training data for video waterdrop removal limits the performance of learning-based methods in the real world. It is almost impossible to collect perfectly aligned driving videos with and without waterdrops.
To address this issue, we propose a large-scale synthetic video waterdrop dataset with the paired data for training. To promote our proposed method to generalize to real driving scenes, we adopt a cross-modality training strategy that jointly trains our model on our large-scale synthetic video dataset and a small-scale real-world image dataset proposed by~\cite{qian2018attentive}. 
We also propose a spatio-temporal fusion-based framework to restore the background information under the regions occupied by sparse waterdrops and even the streaks of waterdrops, as shown in Fig.~\ref{fig:1}.  With the dedicated framework and cross-modality strategy, the proposed method achieves the best-performing video waterdrop removal in real driving scenes.

Our contributions can be summarized as follows.
\begin{enumerate}
    \item We propose a dedicated framework for video waterdrop removal in complex driving scenes. Our framework is based on spatio-temporal fusion that uses a self-attention mechanism to exploit both spatial and temporal information for restoring clean video frames.\\
    \vspace{-2mm}
    \item We are the first to propose a large-scale synthetic video dataset for waterdrop removal task. We design a video waterdrop synthesis algorithm based on Hao~\etal~\cite{hao2019learning} to generate synthetic data.\\
    \vspace{-2mm}
    \item To tackle the domain gap between the synthetic data and real driving scenes, we propose a cross-modality strategy for jointly training the proposed method on our synthetic video data and the real-world image data~\cite{qian2018attentive}.\\
    \vspace{-2mm}
    \item The extensive evaluations demonstrate that our waterdrop removal method significantly outperforms previous works on our synthetic dataset and real driving scenes quantitatively and qualitatively.
\end{enumerate}

\vspace{-2mm}
\section{Related Work}

\subsection{Single-Image Methods}
Although there are many methods focusing on image deraining, they mainly attempt to remove rain streaks instead of waterdrops from images. To remove watedrops from images, some learning-based methods have been proposed in recent years. Eigen~\etal~\cite{eigen2013restoring} propose the first CNN-based method to remove the waterdrops from a single degraded image. Due to the over-shallow network architecture design, their method shows poor removal performance for large and dense waterdrops. Qian~\etal~\cite{qian2018attentive} propose a generative adversarial network (GAN) based method. Their method generates an attention map for each input image. However, with the simple concatenation of the input image and its corresponding attention map, their method still only focuses on the local spatial information instead of the global. Quan~\etal~\cite{quan2019deep} propose a shape-driven attention and channel re-calibration to locate and process waterdrops. However, this method is still limited by the local attention mechanism that does not sufficiently explore the long-range but helpful information from the whole single image. Hao~\etal~\cite{hao2019learning} propose a waterdrop synthesis algorithm for the single image. With their synthetic image dataset, they train a deep network for waterdrop detection and removal. However, their method shows the poor performance in real driving scenes. Quan~\etal~\cite{quan2021removing} propose to utilize the neural architecture search method~\cite{liu2018darts} to generate a complementary network to tackle rain streak removal and waterdrop removal jointly. But their method is still too weak to remove waterdrops from real driving scenes.

\subsection{Multi-Image Methods}
\label{section:related_work_2}
Due to limited clues in a single image for clean background information recovery, some researchers~\cite{you2013adherent,shi2021stereo,liu2020learning,alletto2019adherent} propose to take multiple degraded images as input to reconstruct the clean ones. You~\etal~\cite{you2013adherent} propose to calculate the dense motion change to detect waterdrops in each frame. However, their method needs to retrieve similar but clean information over nearly one hundred frames to restore the regions under waterdrops. Shi~\etal~\cite{shi2021stereo} propose a dedicated framework for the stereo waterdrop removal task. Their method relies heavily on the disparity map in each stereo pair, which is not available in monocular driving videos. A coarse-to-fine method is proposed by Liu~\etal~\cite{liu2020learning} to tackle obstruction removals, such as reflection removal and fence removal. Their method can be extended to waterdrop removal task. Based on the flow estimation, they propose to warp nearby frames to provide helpful information for recovery. Most recently, Alletto~\etal~\cite{alletto2019adherent} propose a spatio-temporal de-raining model to tackle single-image and video waterdrop removal. They also propose a waterdrop synthesis algorithm to simulate driving scenes during rain. Although the abovementioned two methods can remove sparse waterdrops from multiple images, they fail to estimate a correct optical flow when there are numerous waterdrops over a sequence of frames. Considering the bottleneck of flow estimation, we propose to utilize a temporal attention block that directly provides effective information from nearby frames to restore clean background information.

\vspace{-1mm}
\section{Approach}
\vspace{-1mm}
\begin{figure*}[t]
    \centering
    \includegraphics[width=\textwidth]{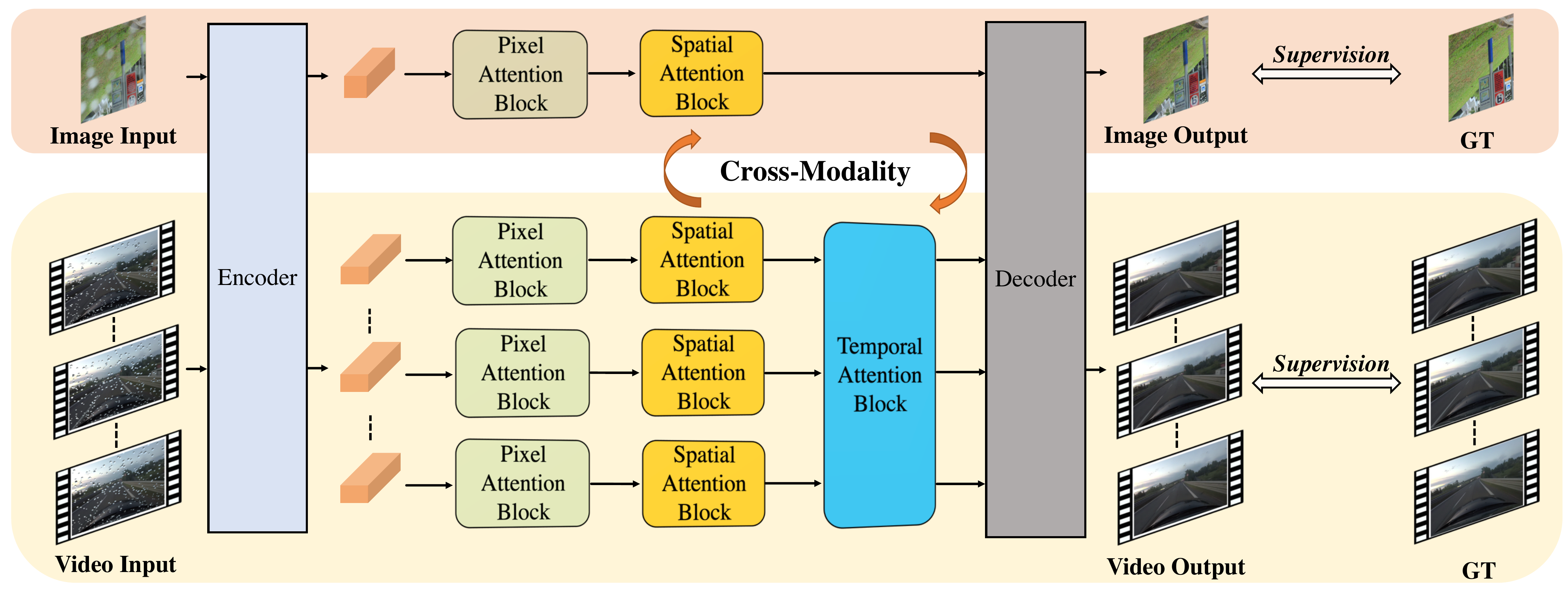}
\vspace{-6mm}    
\caption{Overview of the proposed method. We propose a framework with spatial-temporal fusion, based on a self-attention mechanism \cite{vaswani2017attention}, to reconstruct each feature across spatial and temporal dimensions, respectively. Meanwhile, we also adopt a cross-modality strategy to train our model on our proposed synthetic video data and the real-world image data~\cite{qian2018attentive} jointly.~\label{fig:ill.1}}
\vspace{-6mm}
\end{figure*}

\begin{figure}[t]
    \centering
    \begin{minipage}[t]{0.235\textwidth}
        \centering
        \includegraphics[width=\textwidth]{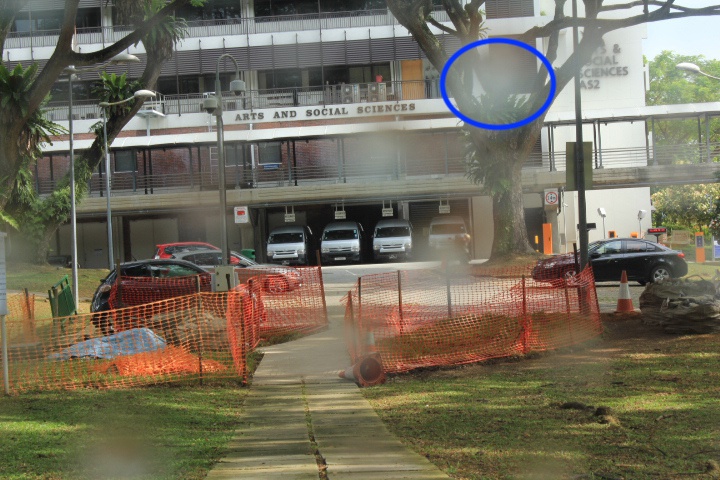}
        \subcaption{Partial occlusion}
    \end{minipage}
    \begin{minipage}[t]{0.235\textwidth}
        \centering
        \includegraphics[width=\textwidth]{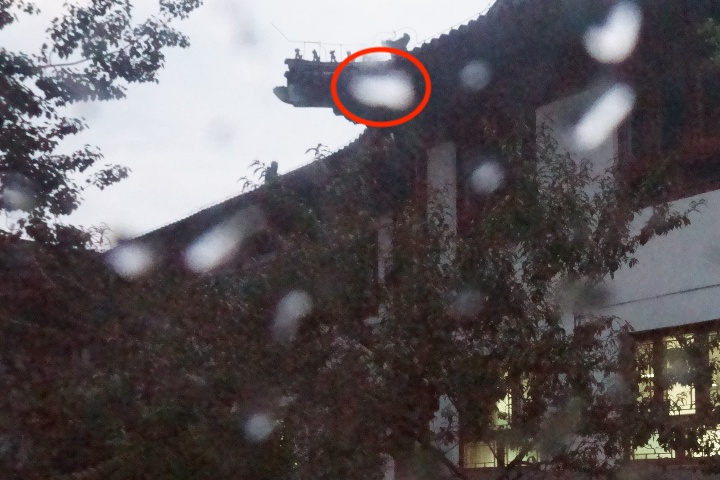}
        \subcaption{Complete occlusion}
    \end{minipage}
\caption{Two types of occlusions caused by waterdrops. Examples are from the dataset~\cite{qian2018attentive}. ~\label{fig:2}}
\vspace{-6mm}
\end{figure}

Given a sequence of frames with waterdrops $\left \{\tilde{F}_{t}\right\}_{t=1}^{T}$, where $T$ is the sequence length, our goal is to remove the waterdrops in these frames and recover the clean ones $\left \{ \hat{F}_{t}\right \}_{t=1}^{T}$. 
To tackle waterdrop removal, we consider two types of occlusions: the partial occlusion and the complete occlusion, as shown in Fig.~\ref{fig:2}. There is still some meaningful background information covered by the waterdrop in the partial occlusion, while the information in the complete occlusion is totally lost for background restoration. To address this problem, we propose a \textit{pixel attention block} to pixel-wisely re-weight the intermediate feature of each input frame. The goal of the pixel attention block is to enhance the background information in the partial occlusion and suppress the meaningless information in the complete occlusion. 

After re-weighting each pixel in the feature, we need to refine the pixels in partially occluded regions and fill completely occluded regions with meaningful values. We treat this processing as an inpainting task~\cite{yu2018generative,zeng2020learning}. Inspired by~\cite{zeng2020learning}, we adopt a self-attention mechanism to refine and fill regions degraded by waterdrops. Unlike previous solely CNN-based works which are limited to exploiting local information, we utilize the self-attention mechanism to restore background information by fusing global information across the whole spatial dimension. Based on such a self-attention mechanism, we propose a \textit{spatial attention block} to refine re-weighted features from the pixel attention block. 

Although the \textit{spatial attention block} can restore most regions in features, some regions that are still degraded can be restored by utilizing features from nearby frames. Hence, we propose a \textit{temporal attention block} to fully exploit the valid information from nearby frames. Similarly, but differently to the self-attention mechanism in the \textit{spatial attention block}, the \textit{temporal attention block} refines multiple features simultaneously by fusing the global information across the temporal dimension. Through the above spatio-temporal fusion, the proposed method can reconstruct cleaned frames accurately.

\subsection{Pixel Attention Block (PAB)}
\begin{figure}[htbp]
    \centering
    \includegraphics[width=0.4\textwidth]{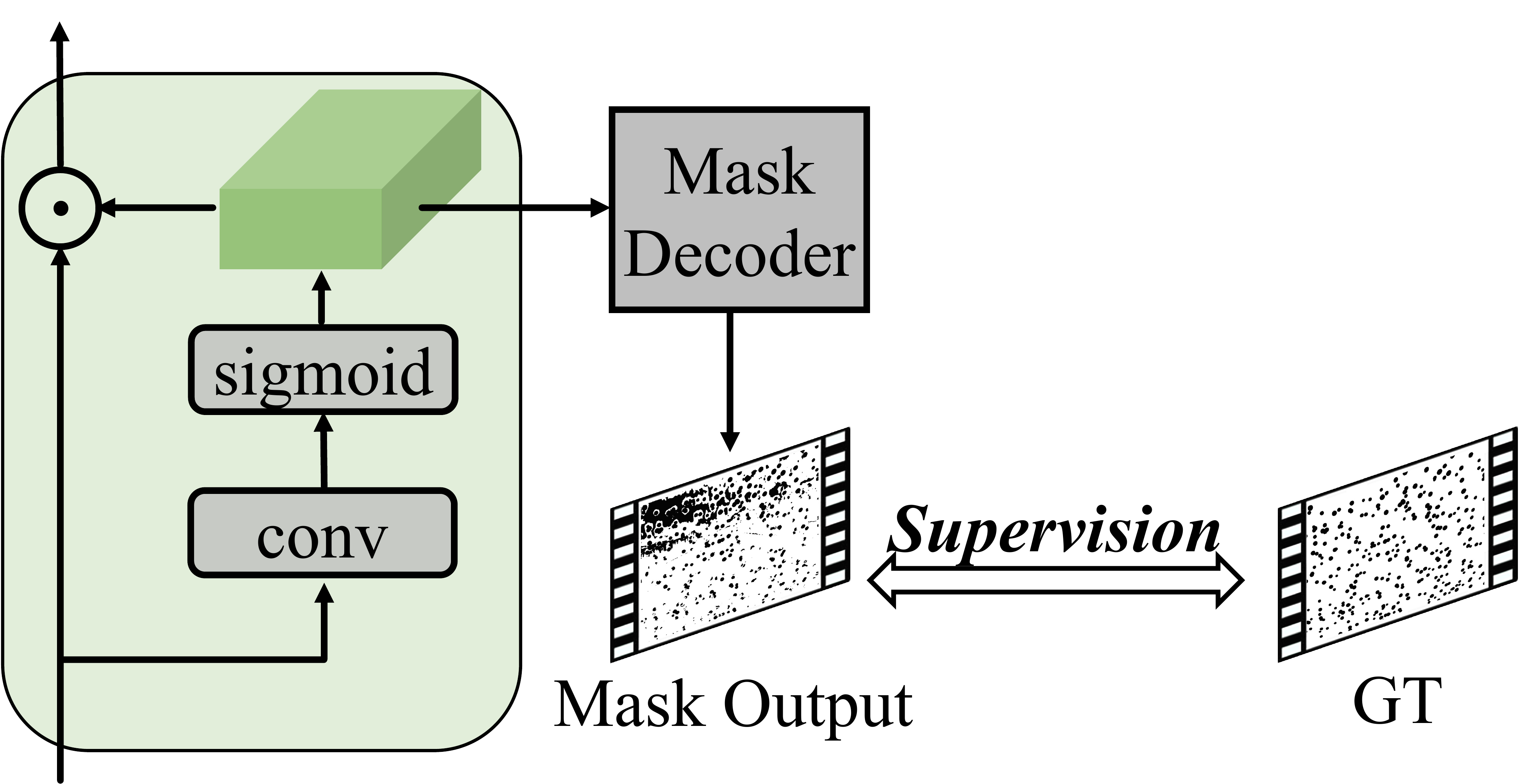}
\vspace{-2mm}
\caption{The detail of the pixel attention block in the proposed method.~\label{fig:ill.2}}
\vspace{-6mm}
\end{figure}

Given a sequence of frames within waterdrops $\left \{\tilde{F}_{t} \right \}_{t=1}^{T}, \tilde{F}_{t}\in \mathbb{R}^{H\times W\times 3}$, we first feed them into the encoder and obtain a sequence of features $\left \{ f_{t} \right \}_{t=1}^{T}, f_{t}\in \mathbb{R}^{h\times w\times c}$, where $h=\frac{H}{4}$, $w=\frac{W}{4}$ and $c$ denotes the channel number. After encoding, each feature is fed into the pixel attention block to obtain its corresponding pixel-wise confidence map $a_t$, as shown in Fig.~\ref{fig:ill.2}:
\begin{equation}
a_t = \sigma(\Theta_{PA}(f_{t})),
\end{equation}
where $1\leq t\leq T$, $\sigma$ denotes the sigmoid activation. To obtain a pixel-wise confidence map for each feature, we adopt a similar block in SENet~\cite{hu2018squeeze} but we discard the global average pooling layer and replace the fully connected layer with the convolutional layer $\Theta_{PA}$. Since each waterdrop only occupies a narrow region across the spatial dimension in each feature channel, this encourages us to re-weight each pixel by a pixel-wise confidence map $a_t$ instead of a common weight as in~\cite{hu2018squeeze}:
\begin{equation}
f^{'}_{t} = a_t\odot f_{t},
\end{equation}
where $\odot$ denotes a pixel-wise multiplication operation.
During re-weighting, the value of each element in the pixel-wise confidence map is expected to be $0$ for the complete occlusion, $1$ for the clean background region.

\subsection{Spatio-Temporal Fusion}
\subsubsection{Spatial Attention Block (SAB)}
After re-weighting, we need to refine the pixels in the partial occlusion and fill the complete occlusion with meaningful pixel values. To address this, we utilize a self-attention mechanism~\cite{vaswani2017attention} to reconstruct each feature by fusing patches with a predicted attention map:
\begin{equation}
\begin{split}
q_{t} = \Psi _{q}(f^{'}_{t}),\quad k_{t} = \Psi _{k}(f^{'}_{t}),\quad v_{t} = \Psi _{v}(f^{'}_{t}),
\end{split}
\end{equation}
where $\Psi _{q}(\cdot)$, $\Psi _{k}(\cdot)$ and $\Psi _{v}(\cdot)$ denote the feature embedding for query, key and value, which consists of the 2D convolutional layers with $1\times 1$ kernel size. \
We extract $N=h \times w$ patches with size $1\times 1\times c$ from the query, key and value features, respectively. To calculate an attention map, we reshape each patch of query and key features into a 1-D vector, so the similarity between each query patch $p^{q}$ and each key patch $p^{k}$ is
\begin{equation}
s_{i,j} = \frac{p^{q}_{i}\cdot (p^{k}_{j})^{T}}{\sqrt{1\times 1\times c}},
\end{equation}
where $1\leq i, j \leq N$, $p^{q}_{i}$ and $p^{k}_{i}$ denote the $i$-th query patch and $j$-th key patch, respectively, $\cdot$ denotes a matrix multiplication operation. As it is mentioned in~\cite{vaswani2017attention}, the similarity value normalized by the dimension of each vector can avoid a small gradient caused by a subsequent softmax function. Finally, the attention map is
\begin{equation}
a_{i,j} = \frac{\mathrm{exp}(s_{i,j})}{\sum_{n=1}^{N}\mathrm{exp}(s_{i,n})}.
\end{equation}
With the attention map, the output $p^{o}_{i}$ for each query patch $p^{q}_{i}$ is a weighted fusion of corresponding value patches $p^{v}$:
\begin{equation}
p^{o}_{i} = \sum_{j=1}^{N}a_{i,j}p^{v}_{j}.
\end{equation}
After receiving these output patches, we reshape them into the size $h\times w\times c$ to obtain the spatially refined feature.

\subsubsection{Temporal Attention Block (TAB)}
In the temporal attention block, we extract multi-scale patches for queries, keys, and values from multiple input frames features as in~\cite{zeng2020learning}. For the waterdrop removal task, the large patches are marvelous for semantic-level reconstruction, and the small ones encourage texture-level reconstruction. For the trade-off between the computation cost and recovering performance, we split each feature into two parts with size $h\times w\times \frac{c}{2}$ and extract different-size patches ($2\times 2\times \frac{c}{2}$ and $8\times 8\times \frac{c}{2}$) from them. After fusing patches with a self-attention mechanism, we reshape fused patches into $h\times w\times c$ features to receive temporally refined features.

After the spatio-temporal fusion, we feed each temporally refined feature into the decoder to obtain cleaned frames $\left \{ \hat{F}_{t} \right \}_{t=1}^{T}$.

\begin{figure*}[t]
    \centering
    \includegraphics[width=\linewidth]{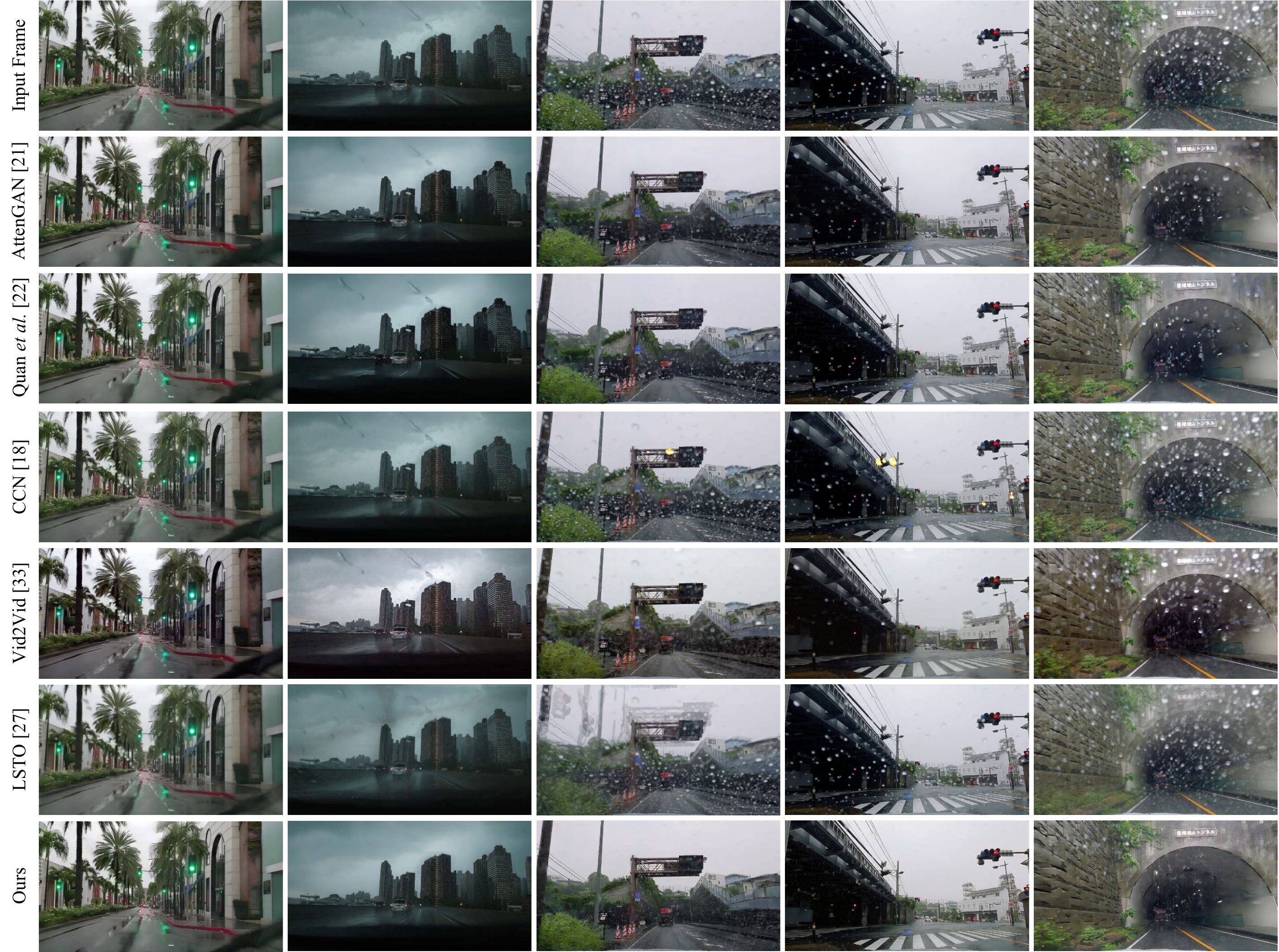}
\vspace{-2mm}
\caption{Qualitative comparison on real driving scenes collected from the Internet. The results show that our method can remove various-shaped waterdrops in multiple kinds of weather. The last column shows the proposed method still presents a satisfying performance in the extremely challenging case.~\label{fig:4}}
\vspace{-7mm}
\end{figure*}

\subsection{Training}
\subsubsection{Cross-modality Training Strategy}
Although the proposed method only trained on our synthetic dataset can generalize well to real driving scenes, training it jointly on our synthetic video data and the real-world image data~\cite{qian2018attentive} as shown in Fig.~\ref{fig:ill.1} improves the generalization performance of our method. As mentioned in~\cite{qian2018attentive}, the dataset~\cite{qian2018attentive} only consists of degraded images with corresponding clean images, and there is a slight spatial misalignment in each image pair, so we need to utilize these image data carefully.

\subsubsection{Loss Functions for Frame Sequences}
\paragraph{Mask Loss} In the pixel attention block, to encourage the network to predict the pixel-wise attention map precisely as much as possible, we feed the predicted pixel-wise attention map into the mask decoder to receive a waterdrop mask for each input frame, the mask loss is
\begin{equation}
\mathcal L^{F}_{mask} = \frac{1}{T}\sum_{t=1}^{T}BCE(\hat{M}_{t}, M_{t}),
\end{equation}
where $BCE(\cdot)$ denotes binary cross entropy loss function, $\hat{M}_{t}$ denotes the predicted mask and $M_{t}$ denotes the ground-truth mask from our synthetic dataset.


\paragraph{Reconstruction Loss} For the final cleaned frame reconstruction, we add a pixel-wise loss between the network outputs and the ground truths:
\begin{eqnarray}
\mathcal L^{F}_{recons} &=& \frac{1}{T}\sum_{t=1}^{T}\left \| \hat{F}_{t} - F_{t} \right \|^{2}_{2},
\end{eqnarray}
where $\tilde{F}_{t}$ denotes the cleaned frame output.\\

\paragraph{Temporal Loss} To guarantee the temporal consistency among network outputs, we adopt the conditional video discriminator \cite{wang2018video} to calculate the temporal loss (TL) for network outputs:
\begin{equation}
\mathcal L^{F}_{TL} = E_{\widehat{x}\sim P_{\mathbf{\tilde{F}_{T}}}(\widehat{x})}[logD_{V}(\widehat{x})],
\end{equation}
where $P(\cdot)$ denotes the data distribution, $\mathbf{\tilde{F}_{T}}$ denotes the concatenation of all the network outputs, $D_{V}$ denotes the discriminator whose loss function is
\begin{equation}
\begin{split}
\mathcal L_{D} = E_{x\sim P_{\mathbf{F_{T}}(x)}}[logD_{V}(x)]+E_{\widehat{x}\sim P_{\mathbf{\tilde{F}_{T}}(\widehat{x})}}[1-logD_{V}(\widehat{x})],
\end{split}
\end{equation}
where $\mathbf{F_{T}}$ denotes the concatenation of all the ground-truth frames from our synthetic dataset.\\

The total loss functions for each frame sequence are concluded as follows:
\begin{equation}
\mathcal L^{F}_{total} = \lambda ^{F}_{1} \cdot \mathcal L^{F}_{mask}++\lambda ^{F}_{2} \cdot \mathcal L^{F}_{recons} +\lambda ^{F}_{3} \cdot \mathcal L^{F}_{TL},
\end{equation}
where the weights for different losses are set as $\lambda ^{F}_{1}=10$, $\lambda ^{F}_{2}=25$, and $\lambda ^{F}_{3}=5$.

\subsubsection{Loss Functions for Images}
Considering the spatial misalignment between the input image $\tilde{I}$ and its clean ground truth $I$, we only adopt a feature matching loss between the network output and the ground truth:
\begin{equation}
\mathcal L^{I}_{feat} =\sum_{l} \lambda _{l}\left \|\Phi_{l}(\hat{I})-\Phi_{l}(I)\right \|_{1},
\end{equation}
where $\hat{I}$ denotes the network output, $\Phi_{l}$ denotes the layer $l$ in the VGG-16 network~\cite{johnson2016perceptual} where we select the layers $\mathrm{conv1}\_2$, $\mathrm{conv2}\_2$, $\mathrm{conv3}\_2$, $\mathrm{conv4}\_2$ and $\mathrm{conv5}\_2$, $\left \{\lambda _{l}\right \}$ denotes the weights for the layers we select. Besides, since there is no need to exploit the temporal information for image reconstruction, we discard the temporal attention block during attention-based fusion.

\vspace{-2mm}
\section{Synthetic Video Data Generation}
Due to the lack of paired data for multi-image/video methods training, we propose a synthetic waterdrop dataset for driving scenes. To the best of our knowledge, there are only three image waterdrop synthesis works~\cite{hao2019learning,alletto2019adherent,porav2019can} which cannot be utilized to generate waterdrops for videos directly. This encourages us to extend such algorithms to video waterdrop synthesis by considering the followings:
\begin{enumerate}
    \item For a driving video, each waterdrop remains at the same position over a sequence of frames. With the wind or the intense movement of the car, there will be a tiny shift in each waterdrop.\\
    \vspace{-2mm}
    \item  With the evaporation of waterdrops, some completely occluded regions turn to be partially occluded along a sequence of frames.
\end{enumerate}
Based on these observations, we design a video waterdrop synthesis algorithm based on~\cite{hao2019learning}. For a sequence of clean frames, we generate 150 to 400 waterdrops in the first frame. For each next frame, we add a shift along a random direction to each waterdrop. Meanwhile, we linearly enlarge the blur kernel size to make waterdrops progressively blurry over frames, which can simulate the evaporation of waterdrops. During synthesizing, we set the length of each sequence as 5, the shift value as 1 pixel, and the blur kernel size from 3 to 20 pixels.

To collect clean driving videos for data synthesis, we choose the DR(eye)VE Dataset~\cite{dreyeve2018} which contains 51 videos with multiple scenes and weathers, varying from the morning, evening, night, sunny, cloudy, countryside, downtown, and highway.

To summarize, we propose a large-scale synthetic dataset with numerous triplets $(F, \tilde{F}, M)$. In each triplet, there are one clean frame $F$, one frame $\tilde{F}$ with synthetic waterdrops and one binary mask $M$ for waterdrops. Totally, we have 67500 triplets from 45 videos for training and 600 triplets from 6 videos for testing.

\vspace{-2mm}
\section{Experiments}
\subsection{Implementation Details}
We train the network by minimizing $(\mathcal L^{F}_{total}+\mathcal L^{I}_{feat})$ on synthetic data and real data jointly. To be specific, for every 1000 training iterations, we train the network on synthetic data for 900 iterations while on real-world data for 100 iterations. We set the proportion as $9:1$, the best one we select from $5:1$, $7:1$, $9:1$ and $11:1$ by evaluations. Totally, we train the network for $75000$ iterations on both datasets using the ADAM optimizer~\cite{DBLP:journals/corr/KingmaB14} with $lr=0.0001$ and $(\beta _{1}, \beta _{2})=(0.9, 0.999)$. We set the length $T$ of each frame sequence as 5, and channel number $c$ as 256 during training.

\vspace{-2mm}
\subsection{Comparison to State-of-the-Art}
As abovementioned in Sec.~\ref{section:related_work_2}, there is only one video waterdrop removal method proposed by Alletto~\etal~\cite{alletto2019adherent}. However, the lack of source code and real-scene evaluations impedes us from making a fair comparison with them. To evaluate our model and make fair comparisons, we select the most recent and the most competitive methods for comparisons:
\begin{enumerate}
    \item AttentGAN~\cite{qian2018attentive}, Quan~\etal~\cite{quan2019deep}, CCN~\cite{quan2021removing} are learning-based methods for single-image waterdrop removal.
    
    \item Vid2Vid~\cite{wang2018video} is a recent and typical video-to-video translation method that is regarded as the spatio-temporal baseline in evaluation.
    
    \item LSTO~\cite{liu2020learning} is a method for multi-image obstruction removal. This method can be extended to video waterdrop removal.
\end{enumerate}

\begin{table}[t!]
    \centering
    \tabcolsep=0.03cm
    \caption{Quantitative comparison on our proposed synthetic dataset. ~\label{table:score}}
    \vspace{-2mm}
    \setlength\arrayrulewidth{1.0pt}
    \resizebox{1\linewidth}{!}{
    \begin{tabular}{cccc}
        \toprule
        \multicolumn{1}{p{1.5cm}}{}
        &\multicolumn{1}{p{2cm}}{\centering CCN~\cite{quan2021removing}}
        &\multicolumn{1}{p{2cm}}{\centering Vid2Vid~\cite{wang2018video}}
        &\multicolumn{1}{p{1.5cm}}{\centering \textbf{Ours}} \\
        \cmidrule(lr){2-4}
        PSNR $\uparrow$ & 26.2878 & 27.5029 & \textbf{29.5789} \\
        MS-SSIM $\uparrow$ & 0.9220 & 0.9439 & \textbf{0.9627} \\
        LPIPS $\downarrow$ & 0.1896 & 0.1231 & \textbf{0.0848} \\
        $E_{warp}$~\cite{lei2020dvp} $\downarrow$ & 0.0828 & 0.0811 & \textbf{0.0799} \\
        \bottomrule
    \end{tabular}}
\vspace{-4mm}
\end{table}

\subsubsection{Qualitative Evaluation}
We present some qualitative comparisons in Fig.~\ref{fig:4}, evaluated on real driving scenes which are collected from the Internet (without ground truth). For various waterdrop cases, image methods show poor performance on video waterdrop removal, especially CCN~\cite{quan2021removing}, since the network architecture dedicated to the single-image task is too weak to handle such complex driving scenes. Although Vid2Vid~\cite{wang2018video} shows satisfying removing performance on the sparse and small waterdrops as shown in Fig.~\ref{fig:4} column 4, this CNN-based method fails to address streak case with large waterdrops and it cannot restore the background information under the complete occlusions, as shown in Fig.~\ref{fig:4} column 1. LSTO~\cite{liu2020learning} can remove most waterdrops in real driving scenes, while this multi-image method relies heavily on flow estimation, which may fail when there are many waterdrops along a sequence of frames. Furthermore, this method is too time-consuming, which costs 60 seconds for each frame of size $960\times 512$, while our method only costs 0.5 seconds for the same one. Based on our pixel attention block and spatio-temporal fusion, the proposed method shows outstanding waterdrop removal performance in sparse cases and streak cases. Even under dark circumstances and heavy rain, the proposed method still presents satisfying results in such extreme cases, as shown in Fig.~\ref{fig:4} columns 2 and 5.

\subsubsection{Quantitative Evaluation}
\paragraph{PSNR and MS-SSIM} Due to the lack of paired data of real driving scenes, we evaluate the different methods on the testing set of our synthetic dataset. Since the lack of training codes for other methods, we only compare the proposed method with CCN~\cite{quan2021removing} and Vid2Vid~\cite{wang2018video}, which are re-trained on the proposed dataset with the same cross-modality training strategy. As shown in Table~\ref{table:score}, we compute the PSNR and MS-SSIM~\cite{wang2003multiscale} between the result cleaned frames of different methods and ground-truth clean frames. The proposed method shows strong performance over previous works on synthetic data.

\paragraph{User Study} We also conduct a user study for the results of different methods on real driving scenes. For each evaluation, we compare our method to the other five methods. Each user is presented with a driving-scene frame degraded by waterdrops and six cleaned frames from different methods. The user needs to choose the cleaned one which has better visual quality. There are 25 driving-scene frames presented in the comparisons and 30 users in this user study. The results are shown in Table \ref{table:user study}. There is nearly $77.84$ percent of users prefer our results, which means our method outperforms others significantly.\\

\begin{table}[t!]
    \centering
    \tabcolsep=0.03cm
    \caption{User study for video waterdrop removal task.~\label{table:user study}}
    \vspace{-2mm}
    {
    \setlength\arrayrulewidth{1.0pt}
    \resizebox{1\linewidth}{!}{
    \begin{tabular}{ccccccc}
        \toprule
        \multicolumn{1}{p{1.5cm}}{\centering AttentGAN\\~\cite{qian2018attentive}}
        &\multicolumn{1}{p{1.5cm}}{\centering Quan~\etal\\~\cite{quan2019deep}}
        &\multicolumn{1}{p{1cm}}{\centering CCN\\~\cite{quan2021removing}}
        &\multicolumn{1}{p{1cm}}{\centering Vid2Vid\\~\cite{wang2018video}}
        &\multicolumn{1}{p{1cm}}{\centering LSTO\\~\cite{liu2020learning}}
        &\multicolumn{1}{p{1cm}}{\centering Ours} \\
        \cmidrule(lr){1-6}
        4.54$\%$ & 10.23$\%$ & 0$\%$ & 6.82$\%$ & 0.56$\%$ & \textbf{77.84$\%$} \\
        \bottomrule
    \end{tabular}}}
\end{table}

\begin{table}[t]
    \centering
    \tabcolsep=0.03cm
    \caption{Ablation study on our proposed synthetic dataset. The best and second-best scores are indicated in \textbf{\textcolor{red}{red}} and \textbf{\textcolor{blue}{blue}}.~\label{table:ablation}}
    \vspace{-2mm}    
    {
    \setlength\arrayrulewidth{1.0pt}
    \resizebox{\linewidth}{!}{
    \begin{tabular}{cccccc}
        \toprule
        \multicolumn{1}{p{1.cm}}{}
        &\multicolumn{1}{p{1.5cm}}{\centering Ours-noPAB}
        &\multicolumn{1}{p{1.5cm}}{\centering Ours-noSAB}
        &\multicolumn{1}{p{1.5cm}}{\centering Ours-noTAB}
        &\multicolumn{1}{p{1.6cm}}{\centering Ours-noCMS}
        &\multicolumn{1}{p{1.4cm}}{\centering \textbf{Ours (full)}}\\
        \cmidrule(lr){2-6}
        PSNR $\uparrow$ & 27.6397 & 28.2355 & 28.0135 & \textbf{\textcolor{red}{30.1052}} & \textbf{\textcolor{blue}{29.5789}} \\
        MS-SSIM $\uparrow$ & 0.9456 & 0.9530 & 0.9492 & \textbf{\textcolor{blue}{0.9625}} & \textbf{\textcolor{red}{0.9627}} \\
        LPIPS $\downarrow$ & 0.1209 & 0.1318 & 0.1363 & \textbf{\textcolor{red}{0.0740}} & \textbf{\textcolor{blue}{0.0848}} \\
        $E_{warp}$~\cite{lei2020dvp} $\downarrow$ & 0.0815 & 0.0810 & 0.0831 & \textbf{\textcolor{blue}{0.0800}} & \textbf{\textcolor{red}{0.0799}} \\
        \bottomrule
    \end{tabular}}}
\vspace{-1mm}
\end{table}

\begin{figure}[!ht]
    \centering
    \begin{minipage}[t]{.15\textwidth}
    \centering
    \includegraphics[width=\textwidth]{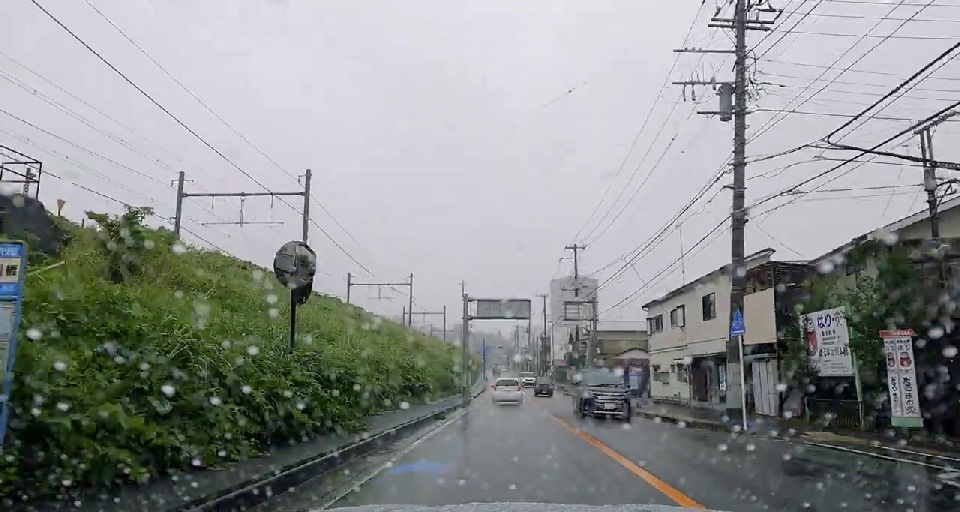}
    \subcaption{Input Frame}
    \end{minipage}
    \begin{minipage}[t]{.15\textwidth}
    \centering
    \includegraphics[width=\textwidth]{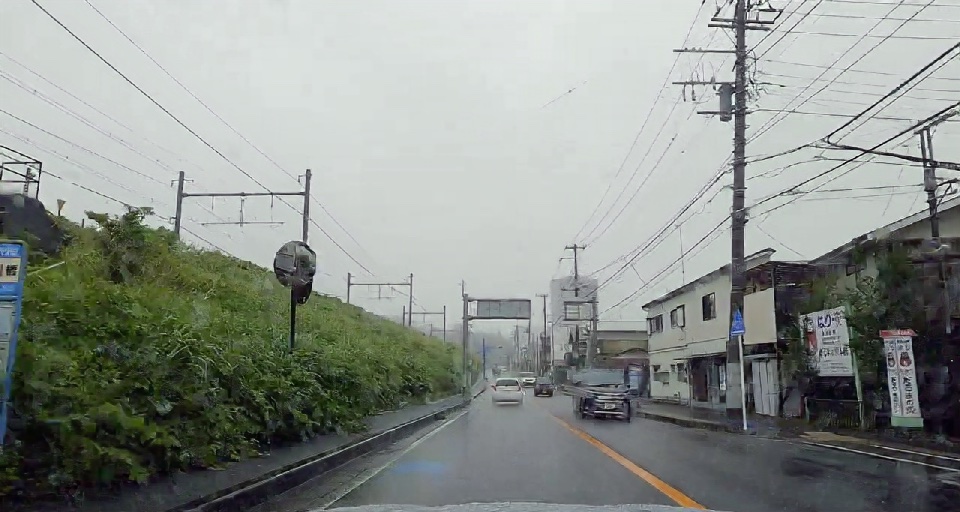}
    \subcaption{Ours-noCMS}
    \end{minipage}
    \begin{minipage}[t]{.15\textwidth}
    \centering
    \includegraphics[width=\textwidth]{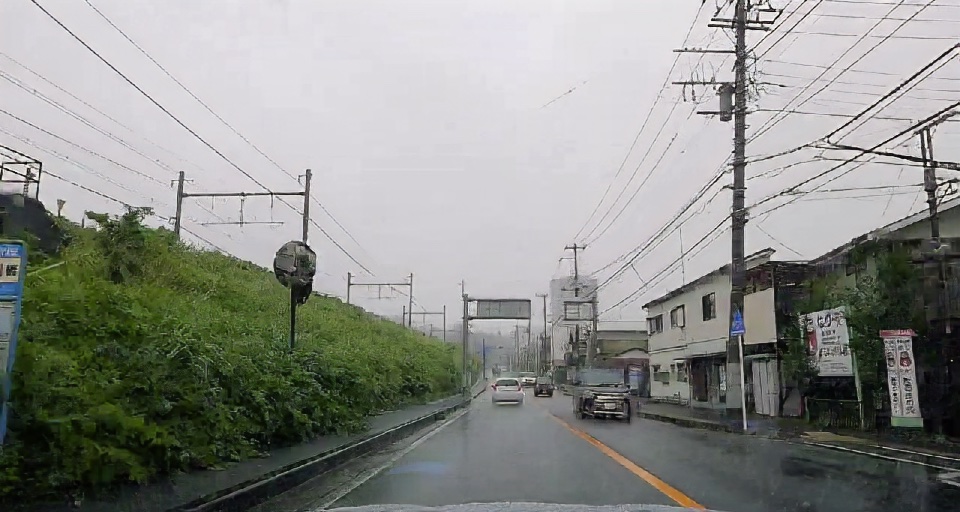}
    \subcaption{Ours (full)}
    \end{minipage}
\caption{Qualitative comparison on real driving scenes for our ablation study.~\label{fig:6}}
\vspace{-4mm}
\end{figure}

\paragraph{Temporal Consistency} To evaluate the quality of results across the temporal dimension, we adopt a similar evaluation metric proposed by Lei~\etal~\cite{lei2020dvp}. For each cleaned frame $\tilde{F}_{t}$, we calculate the warping error with $\tilde{F}_{t-1}$, $\tilde{F}_{t-3}$ and $\tilde{F}_{t-5}$ for considering multi-scale temporal consistency. As shown in Table~\ref{table:ablation}, our model without the temporal attention block suffers heavily from temporal inconsistency, which means the temporal attention block not only restores the regions with the clues from nearby frames but also keeps the network outputs temporally consistent.

\vspace{-2mm}
\subsection{Ablation Study}
To better analyze each block or cross-modality training strategy that contributes to the final performance of our network, we remove or replace each block one by one and re-train each modified network on the same datasets. As shown in Table.~\ref{table:ablation}, each attention block shows a positive effect for removing and recovering performance. Besides, although the model trained without cross-modality strategy (Ours-noCMS), which is only trained on our synthetic dataset, performs the best in quantitative evaluation in Table~\ref{table:ablation}, our full model shows better performance on real driving scenes, as shown in Fig.~\ref{fig:6}. The cleaned frame maintains the original color in good condition with our cross-modality training strategy, and some tiny waterdrops disappear.

\vspace{-1mm}
\section{Conclusion}
\vspace{-1mm}
We present a new method with spatio-temporal fusion for video waterdrop removal. To train our proposed method, we build a large-scale synthetic waterdrop dataset for complex driving scenes. With an appropriate cross-modality training strategy, the proposed method shows an impressive waterdrop removal performance in real driving scenes. Furthermore, we are the first one to compare the proposed method with others on numerous synthetic data and real driving data. Both quantitative and qualitative evaluations show the proposed method is better than others.









{\small
\bibliographystyle{IEEEtran}
\bibliography{ref}

\begin{thebibliography}{10}
\providecommand{\url}[1]{#1}
\csname url@rmstyle\endcsname
\providecommand{\newblock}{\relax}
\providecommand{\bibinfo}[2]{#2}
\providecommand\BIBentrySTDinterwordspacing{\spaceskip=0pt\relax}
\providecommand\BIBentryALTinterwordstretchfactor{4}
\providecommand\BIBentryALTinterwordspacing{\spaceskip=\fontdimen2\font plus
\BIBentryALTinterwordstretchfactor\fontdimen3\font minus
  \fontdimen4\font\relax}
\providecommand\BIBforeignlanguage[2]{{%
\expandafter\ifx\csname l@#1\endcsname\relax
\typeout{** WARNING: IEEEtran.bst: No hyphenation pattern has been}%
\typeout{** loaded for the language `#1'. Using the pattern for}%
\typeout{** the default language instead.}%
\else
\language=\csname l@#1\endcsname
\fi
#2}}

\bibitem{bruls2018mark}
T.~Bruls, W.~Maddern, A.~A. Morye, and P.~Newman, ``Mark yourself: Road marking
  segmentation via weakly-supervised annotations from multimodal data,'' in
  \emph{2018 IEEE International Conference on Robotics and Automation
  (ICRA)}.\hskip 1em plus 0.5em minus 0.4em\relax IEEE, 2018, pp. 1863--1870.

\bibitem{valada2017adapnet}
A.~Valada, J.~Vertens, A.~Dhall, and W.~Burgard, ``Adapnet: Adaptive semantic
  segmentation in adverse environmental conditions,'' in \emph{2017 IEEE
  International Conference on Robotics and Automation (ICRA)}.\hskip 1em plus
  0.5em minus 0.4em\relax IEEE, 2017, pp. 4644--4651.

\bibitem{kumar2021omnidet}
V.~R. Kumar, S.~Yogamani, H.~Rashed, G.~Sitsu, C.~Witt, I.~Leang, S.~Milz, and
  P.~M{\"a}der, ``Omnidet: Surround view cameras based multi-task visual
  perception network for autonomous driving,'' \emph{IEEE Robotics and
  Automation Letters}, vol.~6, no.~2, pp. 2830--2837, 2021.

\bibitem{zou2022real}
Z.~Zou, R.~Zhang, S.~Shen, G.~Pandey, P.~Chakravarty, A.~Parchami, and H.~X.
  Liu, ``Real-time full-stack traffic scene perception for autonomous driving
  with roadside cameras,'' in \emph{2022 International Conference on Robotics
  and Automation (ICRA)}.\hskip 1em plus 0.5em minus 0.4em\relax IEEE, 2022,
  pp. 890--896.

\bibitem{porav2019can}
H.~Porav, T.~Bruls, and P.~Newman, ``I can see clearly now: Image restoration
  via de-raining,'' in \emph{2019 International Conference on Robotics and
  Automation (ICRA)}.\hskip 1em plus 0.5em minus 0.4em\relax IEEE, 2019, pp.
  7087--7093.

\bibitem{fu2017clearing}
X.~Fu, J.~Huang, X.~Ding, Y.~Liao, and J.~Paisley, ``Clearing the skies: A deep
  network architecture for single-image rain removal,'' \emph{IEEE Transactions
  on Image Processing}, vol.~26, no.~6, pp. 2944--2956, 2017.

\bibitem{zhang2019image}
H.~Zhang, V.~Sindagi, and V.~M. Patel, ``Image de-raining using a conditional
  generative adversarial network,'' \emph{IEEE transactions on circuits and
  systems for video technology}, vol.~30, no.~11, pp. 3943--3956, 2019.

\bibitem{chen2021hinet}
L.~Chen, X.~Lu, J.~Zhang, X.~Chu, and C.~Chen, ``Hinet: Half instance
  normalization network for image restoration,'' in \emph{Proceedings of the
  IEEE/CVF Conference on Computer Vision and Pattern Recognition}, 2021, pp.
  182--192.

\bibitem{li2019heavy}
R.~Li, L.-F. Cheong, and R.~T. Tan, ``Heavy rain image restoration: Integrating
  physics model and conditional adversarial learning,'' in \emph{Proceedings of
  the IEEE/CVF Conference on Computer Vision and Pattern Recognition}, 2019,
  pp. 1633--1642.

\bibitem{chen2019gated}
D.~Chen, M.~He, Q.~Fan, J.~Liao, L.~Zhang, D.~Hou, L.~Yuan, and G.~Hua, ``Gated
  context aggregation network for image dehazing and deraining,'' in \emph{2019
  IEEE winter conference on applications of computer vision (WACV)}.\hskip 1em
  plus 0.5em minus 0.4em\relax IEEE, 2019, pp. 1375--1383.

\bibitem{wang2020model}
H.~Wang, Q.~Xie, Q.~Zhao, and D.~Meng, ``A model-driven deep neural network for
  single image rain removal,'' in \emph{Proceedings of the IEEE/CVF Conference
  on Computer Vision and Pattern Recognition}, 2020, pp. 3103--3112.

\bibitem{yi2021structure}
Q.~Yi, J.~Li, Q.~Dai, F.~Fang, G.~Zhang, and T.~Zeng, ``Structure-preserving
  deraining with residue channel prior guidance,'' in \emph{Proceedings of the
  IEEE/CVF International Conference on Computer Vision}, 2021, pp. 4238--4247.

\bibitem{zamir2021multi}
S.~W. Zamir, A.~Arora, S.~Khan, M.~Hayat, F.~S. Khan, M.-H. Yang, and L.~Shao,
  ``Multi-stage progressive image restoration,'' in \emph{Proceedings of the
  IEEE/CVF Conference on Computer Vision and Pattern Recognition}, 2021, pp.
  14\,821--14\,831.

\bibitem{chen2021robust}
C.~Chen and H.~Li, ``Robust representation learning with feedback for single
  image deraining,'' in \emph{Proceedings of the IEEE/CVF Conference on
  Computer Vision and Pattern Recognition}, 2021, pp. 7742--7751.

\bibitem{wang2021rain}
H.~Wang, Z.~Yue, Q.~Xie, Q.~Zhao, Y.~Zheng, and D.~Meng, ``From rain generation
  to rain removal,'' in \emph{Proceedings of the IEEE/CVF Conference on
  Computer Vision and Pattern Recognition}, 2021, pp. 14\,791--14\,801.

\bibitem{chen2020pmhld}
W.-T. Chen, H.-Y. Fang, J.-J. Ding, and S.-Y. Kuo, ``Pmhld: patch map-based
  hybrid learning dehazenet for single image haze removal,'' \emph{IEEE
  Transactions on Image Processing}, vol.~29, pp. 6773--6788, 2020.

\bibitem{jiang2020multi}
K.~Jiang, Z.~Wang, P.~Yi, C.~Chen, B.~Huang, Y.~Luo, J.~Ma, and J.~Jiang,
  ``Multi-scale progressive fusion network for single image deraining,'' in
  \emph{Proceedings of the IEEE/CVF conference on computer vision and pattern
  recognition}, 2020, pp. 8346--8355.

\bibitem{quan2021removing}
R.~Quan, X.~Yu, Y.~Liang, and Y.~Yang, ``Removing raindrops and rain streaks in
  one go,'' in \emph{Proceedings of the IEEE/CVF Conference on Computer Vision
  and Pattern Recognition}, 2021, pp. 9147--9156.

\bibitem{Zamir2021Restormer}
S.~W. Zamir, A.~Arora, S.~Khan, M.~Hayat, F.~S. Khan, and M.-H. Yang,
  ``Restormer: Efficient transformer for high-resolution image restoration,''
  in \emph{CVPR}, 2022.

\bibitem{eigen2013restoring}
D.~Eigen, D.~Krishnan, and R.~Fergus, ``Restoring an image taken through a
  window covered with dirt or rain,'' in \emph{Proceedings of the IEEE
  international conference on computer vision}, 2013, pp. 633--640.

\bibitem{qian2018attentive}
R.~Qian, R.~T. Tan, W.~Yang, J.~Su, and J.~Liu, ``Attentive generative
  adversarial network for raindrop removal from a single image,'' in
  \emph{Proceedings of the IEEE conference on computer vision and pattern
  recognition}, 2018, pp. 2482--2491.

\bibitem{quan2019deep}
Y.~Quan, S.~Deng, Y.~Chen, and H.~Ji, ``Deep learning for seeing through window
  with raindrops,'' in \emph{Proceedings of the IEEE/CVF International
  Conference on Computer Vision}, 2019, pp. 2463--2471.

\bibitem{shi2021stereo}
Z.~Shi, N.~Fan, D.-Y. Yeung, and Q.~Chen, ``Stereo waterdrop removal with
  row-wise dilated attention,'' \emph{IROS}, 2021.

\bibitem{hao2019learning}
Z.~Hao, S.~You, Y.~Li, K.~Li, and F.~Lu, ``Learning from synthetic
  photorealistic raindrop for single image raindrop removal,'' in
  \emph{Proceedings of the IEEE/CVF International Conference on Computer Vision
  Workshops}, 2019, pp. 0--0.

\bibitem{liu2018darts}
H.~Liu, K.~Simonyan, and Y.~Yang, ``Darts: Differentiable architecture
  search,'' in \emph{International Conference on Learning Representations},
  2018.

\bibitem{you2013adherent}
S.~You, R.~T. Tan, R.~Kawakami, and K.~Ikeuchi, ``Adherent raindrop detection
  and removal in video,'' in \emph{Proceedings of the IEEE Conference on
  Computer Vision and Pattern Recognition}, 2013, pp. 1035--1042.

\bibitem{liu2020learning}
Y.-L. Liu, W.-S. Lai, M.-H. Yang, Y.-Y. Chuang, and J.-B. Huang, ``Learning to
  see through obstructions,'' in \emph{Proceedings of the IEEE/CVF Conference
  on Computer Vision and Pattern Recognition}, 2020, pp. 14\,215--14\,224.

\bibitem{alletto2019adherent}
S.~Alletto, C.~Carlin, L.~Rigazio, Y.~Ishii, and S.~Tsukizawa, ``Adherent
  raindrop removal with self-supervised attention maps and spatio-temporal
  generative adversarial networks,'' in \emph{Proceedings of the IEEE/CVF
  International Conference on Computer Vision Workshops}, 2019, pp. 0--0.

\bibitem{vaswani2017attention}
A.~Vaswani, N.~Shazeer, N.~Parmar, J.~Uszkoreit, L.~Jones, A.~N. Gomez,
  {\L}.~Kaiser, and I.~Polosukhin, ``Attention is all you need,'' in
  \emph{Advances in neural information processing systems}, 2017, pp.
  5998--6008.

\bibitem{yu2018generative}
J.~Yu, Z.~Lin, J.~Yang, X.~Shen, X.~Lu, and T.~S. Huang, ``Generative image
  inpainting with contextual attention,'' \emph{arXiv preprint
  arXiv:1801.07892}, 2018.

\bibitem{zeng2020learning}
Y.~Zeng, J.~Fu, and H.~Chao, ``Learning joint spatial-temporal transformations
  for video inpainting,'' in \emph{European Conference on Computer Vision},
  2020, pp. 528--543.

\bibitem{hu2018squeeze}
J.~Hu, L.~Shen, and G.~Sun, ``Squeeze-and-excitation networks,'' in
  \emph{Proceedings of the IEEE conference on computer vision and pattern
  recognition}, 2018, pp. 7132--7141.

\bibitem{wang2018video}
T.-C. Wang, M.-Y. Liu, J.-Y. Zhu, N.~Yakovenko, A.~Tao, J.~Kautz, and
  B.~Catanzaro, ``Video-to-video synthesis,'' in \emph{NeurIPS}, 2018.

\bibitem{johnson2016perceptual}
J.~Johnson, A.~Alahi, and L.~Fei-Fei, ``Perceptual losses for real-time style
  transfer and super-resolution,'' in \emph{European conference on computer
  vision}.\hskip 1em plus 0.5em minus 0.4em\relax Springer, 2016, pp. 694--711.

\bibitem{dreyeve2018}
A.~Palazzi, D.~Abati, S.~Calderara, F.~Solera, and R.~Cucchiara, ``Predicting
  the driver's focus of attention: the dr(eye)ve project,'' \emph{IEEE
  Transactions on Pattern Analysis and Machine Intelligence}, 2018.

\bibitem{DBLP:journals/corr/KingmaB14}
\BIBentryALTinterwordspacing
D.~P. Kingma and J.~Ba, ``Adam: {A} method for stochastic optimization,'' in
  \emph{3rd International Conference on Learning Representations, {ICLR} 2015,
  San Diego, CA, USA, May 7-9, 2015, Conference Track Proceedings}, Y.~Bengio
  and Y.~LeCun, Eds., 2015. [Online]. Available:
  \url{http://arxiv.org/abs/1412.6980}
\BIBentrySTDinterwordspacing

\bibitem{lei2020dvp}
C.~Lei, Y.~Xing, and Q.~Chen, ``Blind video temporal consistency via deep video
  prior,'' in \emph{Advances in Neural Information Processing Systems}, 2020.

\bibitem{wang2003multiscale}
Z.~Wang, E.~P. Simoncelli, and A.~C. Bovik, ``Multiscale structural similarity
  for image quality assessment,'' in \emph{The Thrity-Seventh Asilomar
  Conference on Signals, Systems \& Computers, 2003}, vol.~2.\hskip 1em plus
  0.5em minus 0.4em\relax Ieee, 2003, pp. 1398--1402.

\end{thebibliography}
}

\end{document}